# AI-Guided Feature Segmentation Techniques to Model Features from Single Crystal Diamond Growth

Rohan Reddy Mekala[1,a], Elias Garratt[2,b], Matthias Muehle[3,c], Arjun Srinivasan[1,d], Adam Porter[4,e] and Mikael Lindvall[1,f]

[1]Fraunhofer USA, Center Mid-Atlantic, US

[2]Michigan State University, US

[3]Fraunhofer USA, Center Midwest, US

[4]Fraunhofer USA, Center Mid-Atlantic & Department of Computer Science, University of Maryland, US

[a]rreddy@fraunhofer.org, [b]garrate@msu.edu,
[c]mmuehle@fraunhofer.org, [d]asrinivasan@fraunhofer.org, [e]aporter@fraunhofer.org,
[f]mikael@fraunhofer.org

**Abstract.** Process refinement to consistently produce high quality material over a large area of the grown crystal, which would enable a variety of applications from optics crystals to quantum detectors, has been a long-time goal for diamond growth. Machine learning offers a high value path forward to this goal, but model development comes with several challenges. Primarily these are: the complexity of features within datasets and their time-dependency, and the volume of data produced per growth run. Accurate spatial feature extraction from image to image for real- time monitoring of diamond growth is a crucial yet complicated problem due to the low-volume and high feature complexity nature of the datasets. Additionally, given the complex feature characteristics within images captured in terms shape, size etc., ensuring data consistency, accuracy and integrity as part of the extraction algorithm is a significant challenge. This paper compares various traditional and machine learning driven approaches for feature extraction in the diamond growth domain, and proposes a novel deep learning driven semantic segmentation approach to isolate and classify accurate pixel masks of geometric features like diamond, pocket holder and background, and their corresponding derivative features based on the shape and size. Using an annotation focused human-in-the- loop software architecture to produce training datasets, with modules for selective data labeling using active learning, data augmentations and model assisted labeling, our approach achieves effective annotation accuracy and drastically reduces the time and cost of labeling by several orders of magnitude. In our model development, we found deep learning algorithms to be highly efficient in accurately learning complex representations from datasets with many features. Our top-performing model, based on the DeeplabV3plus architecture, achieved outstanding accuracy in classifying features of interest. Specifically, it achieved accuracies of 96.31% for pocket holder, 98.60% for diamond top, and 91.64% for diamond side features.

## Introduction

The ability to create large single crystal diamonds at high quality (low dislocation density, high chemical purity, and smooth surfaces) is a longstanding and very high value challenge for the scientific community, and one that has yet to be surmounted despite over decades of sustained effort. Moreover, the ability to produce such diamonds at scale represents a significant future challenge for manufacturers of electronic- grade or quantum grade diamonds.

Current diamond material process development relies on reactive approaches, where human operators design experiments based on post-process analysis. However, process conditions monitored during growth are not utilized for predictive analysis due to the complexity of factors involved, such as temperature changes, gas composition, and diamond expansion. To achieve desired outcomes, these factors must be considered collectively during the growth process. The size and shape of a growing diamond reflects the convergence of various factors. Specifically, the morphology and geometric shape directly correlate with growth conditions. A rounded shape suggests deviations from ideal parameters, prompting necessary adjustments. Assessing the diamond's growth quality through in-situ imaging facilitates predictive diagnostics of the growth process.

To automate proactive learning in diamond growth, AI algorithms are essential for uncovering nonlinear correlations between qualitative growth aspects and spatial/numerical parameters. These algorithms, particularly those in computer vision and deep learning, excel in feature extraction and prediction from image sequences. In this paper we make use of ML/DL algorithms [1] to extract and analyze spatial macro- features from in-situ growth data, aiming to guide diamond growth towards optimal outcomes like high-quality CVD diamond plates. These AI-guided approaches promise shortened development cycles by leveraging vast in-situ data, enabling predictive adjustments of process parameters. By training models on extracted features, we can track growth progress and optimize outcomes based on shape and morphology, reducing the need for reactive experimentation and allowing continual updates as conditions evolve. This advancement in deep learning enables the realization of such proactive learning pipelines for diamond growth.

**Contributions**

This paper outlines research and development of a novel pipeline based on DL-driven semantic segmentation algorithms to extract independent and derivative features of interest from time-sequenced images taken during diamond growth. Through the development of the proposed DL-backed pipeline, we advance the state-of-the-art feature extraction for diamond growth in the following ways.
1. This is a novel attempt at the successful development of a DL-based pipeline for feature extraction in the diamond growth domain for low-volume high-feature- complexity training dataset environments where both data procurement and data annotation are extremely expensive tasks. That is, our approach is designed with the time and expense of growing diamond in CVD reactors and complex features that are generated during growth, like diamond shape. Additionally, our approach achieves excellent segmentation accuracy metrics and serves as the first benchmark established for automated feature extraction in the domain of diamond growth.
2. This is a novel crowd-backed labeling pipeline designed, implemented, and validated for creating feature-annotated image datasets for semantic segmentation in diamond growth. Our implementation significantly reduces labeling time and costs (from 13 minutes to 3.2 minutes per training image), ensuring consistency, accuracy, and integrity of labeled data.
3. Moreover, our DL-based pipeline, using our best performing model, achieved excellent feature level Intersection over Union (IoU) accuracy [2] of 96.31%, 98.60% and 91.64% for the pocket holder, diamond top and diamond side features of interest (FOI) respectively.
4. This is a novel attempt at determining and comparing accuracy measures for diamond growth feature extraction objectives against a near-exhaustive range of state-of-the-art DL models.
5. This paper extends previous objectives by evaluating model permutations based on variations in input image resolution and dataset size. It showcases insights gained from experiments to enhance output precision of feature extraction models, demonstrating the potential for improved accuracy metrics with increased image resolution and training data volume for specific model architectures.

Our research is the first step towards deep learning as a feature extraction mechanism in guiding the automated analysis of diamond synthesis pipelines. The benefit of high detection accuracy, and extremely low compute times, without relying on voluminous labeled data for the training sets should serve as a foundation to further develop the design and methods proposed in the paper for better feature extraction benchmarks in the diamond growth domain.

**Related Work**

The field of feature extraction, while novel to the domain of diamond growth in general, has been previously tackled for sub-domains within manufacturing like additive manufacturing. These implementations used to solve feature extraction objectives have traditionally been employed across the categories described in the following sub section.

**Computer Vision and Machine Learning based Methods in Material Synthesis.** Computer vision algorithms have been traditionally used in the manufacturing niche to solve objectives like shape detection and image classification. For instance, a simple computer vision-based pipeline [3] was used in monitoring the manufacturing of electro-mechanical parts such as wires and switches using traditional image processing techniques. These methods require time-extensive manual feature engineering and can only handle basic features. Statistical machine learning algorithms such as Gaussian process-based predictive models support vector machines, K- nearest neighbors (KNN), and principal component analysis (PCA) [4] have been used in the manufacturing domain in the past for quality monitoring and improvement. As with statistical machine learning techniques, these algorithms only work well on less complex input features and often fail for complex image recognition tasks involving instance semantic segmentation [5].

**Deep Learning Methods in Material Synthesis.** Deep learning (DL) based methods renowned for their exceptional performance in state-of-the-art accuracy metrics have been employed in the past for applications involving automated inspection and quality monitoring. For instance, DL has been used for classifying the microstructures of steel [6] and segmenting out the surface defects from the manufactured materials [7]. A significant bottleneck in these implementations is the need for extensive labeled datasets, which can be costly and time-consuming to acquire. In our research, we address this issue with a DL-based pipeline tailored for low-volume training datasets, achieving excellent segmentation accuracy.

**Background**

AI algorithms aim to expedite diamond growth process development by transforming traditional human-driven post-process analysis and adjustment to machine-guided during-process analysis, prediction, and adjustment. Unlike reactive- experiments, which yield a single outcome, guided approaches continuously monitor real-time data and preemptively adjust process conditions, setting forth new experiments as needed based on evolving product states and emergent data points.

Using in-situ image data for training poses challenges due to feature variability and the numerous parameters affecting future diamond states. High-dimensional spatio-temporal data prediction for images, a cutting-edge field in AI research, faces unsolved challenges like checkerboard artifacts and feature loss. Merging numerical input parameters with sequential images adds complexity. Our work demonstrates proof-of-concept solutions for growth state prediction with high accuracy, addressing these challenges. Automated contour identification is crucial for measuring feature property changes across sequential image states. The overall pipeline envisioned for the diamond growth monitoring is organized into three separate thrusts to demonstrate the applicability of AI models towards exhaustively understanding diamond growth on a predictive and prescriptive level.

1. Feature Extraction: Developing object detection and segmentation models to classify and isolate accurate pixel masks of features like the diamond, pocket holder and background, and their corresponding derivative features.
2. Defect Extraction: Developing object detection and segmentation models to classify and isolate accurate pixel masks of defects like polycrystalline growth, center defects and edge defects, and their corresponding derivative features.
3. Frame Prediction: Developing accurate models of micron-scale growth in single crystal diamonds using reactor input parameters across sequential time states to accurately predict future states including diamond/pocket-holder shapes and macroscopic defects.

In this paper, we focus on research conducted towards the first thrust of feature extraction to extract general entities of interest from reactor images taken in-situ during the diamond growth process. We will provide details on feature extraction pipelines developed using techniques ranging from traditional computer vision techniques to advanced deep learning-based methods and establish benchmarks on accuracy metrics achieved on a general and feature-wise level.

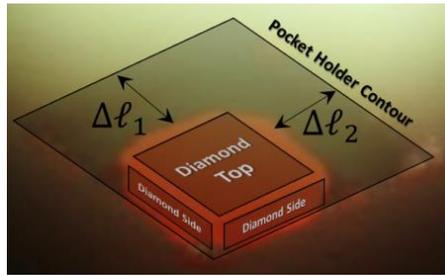
Figure 1. Independent and Derived FOIs for the segmentation pipeline

**Features of interest in Modelling Diamond Growth.** On a broad level, our feature extraction pipeline focuses on extracting the following FOI groups illustrated in Figure 1. The Independant features of interest include:

1. Diamond Top refers to the upper surface of the diamond contour, crucial for monitoring diamond growth as it evolves in shape and size [8]. The morphology of this surface, including (100), (111), or (110) surfaces, is indicative of the diamond's quality under optimal growth conditions. For example, an octahedral shape may form from a square shape for the (100) surface. Variations such as circular shapes suggest excess heat, leading to roughening and non-octagonal shapes [9]. The contours of the top facet provide insight into the growth conditions, particularly at the edges, and allow measurement of the lateral growth rate. This evaluation links the lateral growth rate to top surface morphology and pocket dimensions.

2. Diamond Side comprises the sides of the diamond contour. In optical images, they are typically separated from diamond top by a systematic drop in pixel brightness. These contours are selected to distinguish them from the top surface contour particularly as they do not grow outward and contribute to the top surface shape.

3. Pocket Holder is the recessed area housing the diamond during growth. Its dimensions impact the size the diamond can reach. Over time, polycrystalline diamond (PCD) forms on the holder's edges, reducing available space for diamond growth. This changing environment affects heat distribution and growth conditions. Monitoring the pocket holder's contours helps track these changes and correlates them with diamond quality evolution.

The derived features of interest comprise FOIs obtained by performing operations on the direct output of our feature extraction models to obtain meta information over the FOIs which include:

1. Area of the diamond and pocket holder- This is calculated by aggregating the number of pixels for the diamond and pocket holder contours and serves as a valuable indicator in assessing the rate of increase of diamond area over time.

2. Gap between the pocket holder and diamond contour} - The geometries of the diamond and pocket holder setup is used to calculate the gap between them based on their corresponding mask contours. This is an indicator for diamond crystal growth across the lateral and longitudinal dimensions.

3. Shape of the Diamond crystal- This is used to understand the shape of diamond crystals formed which can be square or octahedral depending on the growth run type. Image processing techniques like contour and edge detection are used to obtain the shape from the diamond-top mask.

**Image Recognition Task of Interest.** This paper explores three classes of techniques for FOI extraction: Object Detection, which predicts bounding boxes for object classes; Semantic Segmentation, which classifies each pixel into multiple classes without distinguishing between instances; and Instance Segmentation, which also distinguishes between instances of the same class at the pixel level. High precision and accuracy in measuring changes in independent FOI contour areas over time are crucial for automated calculations. Object detection algorithms are ineffective as they lack pixel-level precision needed for FOI contour identification. Semantic segmentation is chosen for its pixel-level classification capabilities, especially considering the limited increase in FOI pixels over time in diamond growth. In subsequent sections, we discuss algorithms tested for semantic segmentation model development.1. Traditional Computer Vision methods relying on color/region-based thresholding and edge detection, necessitate manual tuning and are insufficient for complex features in diamond images. 2. Statistical machine learning methods like k-Means clustering and superpixel segmentation cluster pixels based on color intensities and spatial arrangements but require human intervention for fine-tuning, making them unsuitable for complex feature distributions in diamond growth analysis. Due to extreme spatial variations and complexities in diamond growth, traditional algorithms are impractical and require extensive manual tuning. Neural networks/deep learning offer nonlinear learning capabilities better suited for capturing complex distribution patterns, as detailed in subsequent sections. 3. Deep learning semantic segmentation algorithms have been crucial in our research for segmenting FOIs in the diamond growth domain. These algorithms typically utilize convolutional neural networks, multi-scale and pyramid networks, and encoder-decoder architectures. In this paper, we experimented with the state-of-the-art model architectures along with fine- tuning their architectures, training, and deployment for semantic segmentation. A) The Fully Convolutional Network (FCN) architecture [10] consists of an encoder-decoder combination. The encoder downsamples the image to learn a feature vector, while the decoder reconstructs segmentation masks by up sampling and fusing feature maps from previous pooling layers. We found the 8x-up sampling variant most effective for diamond growth (further discussed in next section). The implementation's encoder utilizes the MobileNet-V2 architecture, employing depth wise separable convolution layers for computational efficiency and reduced overfitting risks. B) DeeplabV3 The semantic segmentation model architecture [11] employs deep convolutional neural networks for feature extraction, spatial pyramid pooling for context extraction at various scales, and 16x-upsampling for FOI mask retrieval. DeepLabV3 stands out for its use of atrous convolutions and atrous spatial pyramid pooling (ASPP). Atrous convolution enables controlling the effective field-of-view without impacting computation time, while ASPP captures objects at multiple scales. These atrous layers address information loss caused by feature map size reduction in deep convolutional and pooling layers, common in FCN architectures. C) DeeplabV3+ model architecture [12] is an improvisation over the DeeplabV3 model, in using a separate decoder module to recover object boundaries. Additionally, it uses dilated depthwise separable convolutions in its encoder and decoder modules. It employs DeeplabV3 before upsampling and uses the combined network as the encoder module. The Xception13 architecture employs depthwise separable convolution and is used as backbone in the encoder mod- ule, thereby reducing computations without degrading performance and accuracy. In our implementation, the encoded features are first 4x-upsampled and then concatenated with the corresponding low-level features from the network backbone. A $1 \times 1$ convolution is applied on the low-level features be- fore concatenation to reduce the number of channels and to give more importance to the encoded features. After concatenation, a $3 \times 3$ convolution is performed to refine the features followed by a 4x-upsampling sub layer to recover the FOI masks.

**Method And Design of Experiments**

We devised an AI-powered pipeline to extract features from time-sequenced reactor images, enabling post-analytics on diamond growth. This pipeline aims to achieve state-of-the-art segmentation accuracy for all independent FOIs despite limited training data. Models developed require labeled images with semantic masks encoding FOI contours. Generalization on unseen data depends on diverse training images representative of real-world scenarios, but obtaining such data is costly and time-consuming. Accurate labeling is challenging due to complex FOI characteristics. Thus, the training pipeline incorporates intelligent data selection, augmentation, and selective labeling with feedback loops involving manufacturers, labelers, and models.
Subsequent sections, or modules provide a detailed explanation of the proposed development pipeline. They include descriptions of custom modules for data collection and pre-processing, model research and development (R&D), evaluation, and post analytics.
**Growth Run Data Procurement Experiment Module, SP1.** As part of the data procurement module, we collected images for 25 growth runs in the RAW format using a full-frame mirrorless interchangeable lens camera (Sony Alpha 7R IV), equipped with a macro lens (Sony SEL90M28G) via an optical view port of the reactor's electromagnetic cavity. On an average, data from each growth run procured spans ~34 hours of image and reactor telemetry data (i.e., substrate temperature, reactor pressure, gas flow rates). Images were collected at a 1-minute frequency, while reactor telemetry data was logged every second. Data was synced to a remote storage server for co-ordination between the manufacturing and data science team. Reactor configuration and conditions were kept within previously mapped, stable growth regimes [14].

**Data Pre-processing Experiment Module, SP2.** The data pre-processing module to process our image datasets is developed to produce images ready for consumption by the labeling/annotation and machine learning model development modules in two stages. As part of the first stage to process data to be ready for annotation, the time sequenced image data is re-sampled over 15-minute windows and corresponding images converted to the PNG format to ensure minimal information loss. The selected dataset is consequently run through additional sanity checks using traditional computer vision techniques to filter out noisy or blacked out images created due to intermittent hardware issues. Around 892 images were selected, and the processed dataset is made available to the labeling module, SP3 for processing. While the dataset volume can both be considered small, we were still able to achieve state of the art accuracy metrics by developing a robust framework comprising custom sub-modules for active-learning governed data selection, augmentation and annotations, and model development. The sub-modules will be explained in detail in subsequent sections.

For the second stage of data pre-processing, the datasets are further fine-tuned for consumption by the machine learning model development module, SP4. As part of this stage, the images were first cropped to a smaller bounding box containing all independent FOI contours and resized to two groups of 256x256 and 512x512 resolutions. Subsequently, the images are de-noised [15] and normalized by subtracting all the pixels by 127.5, and then dividing them by 127.5 to map them to a [-1,1] range. Following the pre-processing, to create the target dataset, we randomized the image datasets and split them in a ratio of 90:10 to produce the training and test datasets. The training dataset, once annotated, was used to develop and train the ML models, and the test dataset to evaluate the fit of the trained ML models over various predefined performance metrics. For larger datasets, a greater train to test ratios is likely better, but this ratio helped us ensure maximal dataset size for training/learning, which achieved a slightly better result than an 80:20 split. A larger dataset could also facilitate experiments with various cross-validation techniques to extract generalizable characteristics from our trained models.

**Data Labeling Experiment Module, SP3.** The AI-based human in the loop data labeling module to develop annotations for FOIs within the image datasets is a critical component of our overall pipeline and pivotal in solving for the issues concerning low volume datasets and the expensive nature of procuring both the images and developing the corresponding annotations in a fast and accurate manner. Using the module developed, we were able to reduce the average labeling time for each image from ~13 minutes to ~3.2 minutes.

The module sub-pipeline, as portrayed in Figure 2, starts with saving the pre-processed image data from the SP1/SP2 modules as the input image database for use in annotation development. On average, when using the labeling platform in isolation, labeling each image accurately for the geometry-based semantic segmentation objective takes ~13 minutes. Given this time-intensive nature of annotating images, we customize selection of images using a selection algorithm based on active-learning [16] to only label a subset of the images that potentially increase the segmentation model accuracy metrics towards corresponding FOIs. The selected images are then passed on in batches of 100 images each to the labeling process.

For the consequent labeling process, we use a crowdsourced labeling platform for image recognition tasks called LabelBox [17] to label the growth run images. The first subset of 100 images involves a team of 3 material scientists and 15 external labelers from Labelbox's crowdsourced team. Material scientists provide detailed instructions via videos and meetings, with iterative exchanges continuing until completion. Annotations undergo crowdsourced peer review, with predominant occurrences earmarked based on consensus scores, followed by a final review by material scientists for accuracy and consistency.

Difficulties arise in drafting common FOI annotation instructions due to complex edge cases and varying interpretations among labelers. To address this, we implement model-assisted labeling (MAL) [18], where an incrementally trained model overlays pre-filled contour outputs onto images for labelers to refine. This approach significantly reduces labeling time from ~13 minutes to ~3.2 minutes and enhances label consistency and accuracy. Further details on model architecture development are discussed in the next section Model Research and Development Experiments, SP4. The MAL process continues until the model achieves a segmentation accuracy threshold of 80%, after which future batches use contour overlays from the finalized baseline model for annotation.

Once the final set of images-label pairs meet the requisite requirements on the minimum number of images processed, they are passed over for use in research and development of the final version of the semantic segmentation models, as explained in the next module.

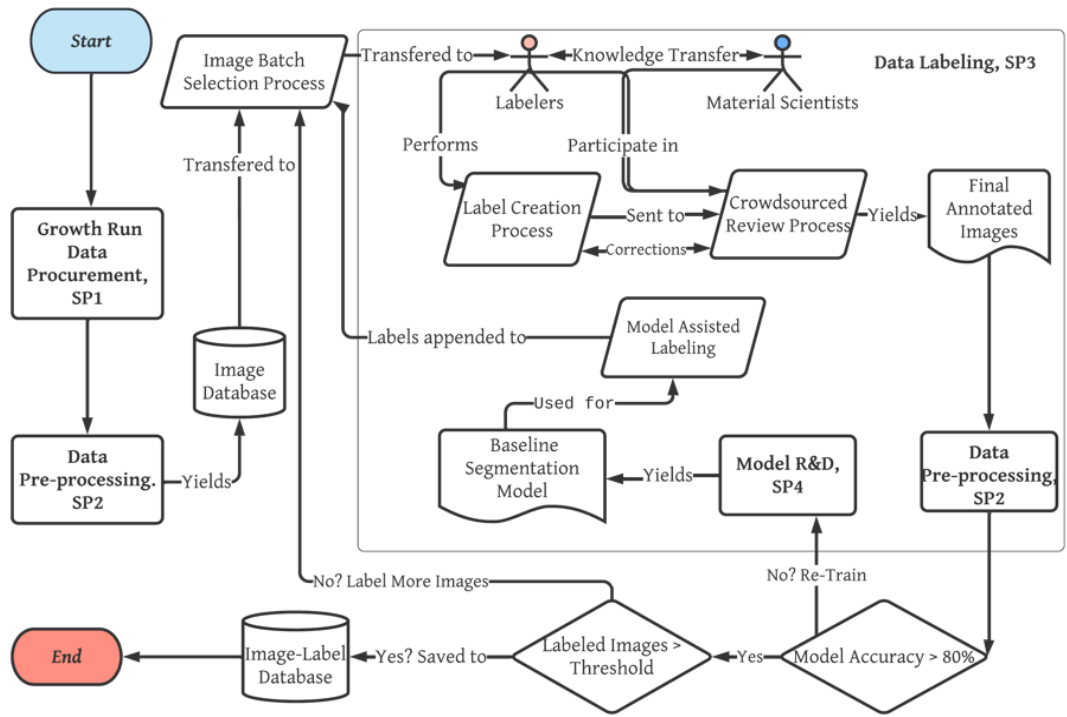

Figure 2. Flow chart explaining the Data Development and Labeling Modules

**Model Research and Development Experiments, SP4.** The model research and development module comprise the use of image-label datasets from SP3 for development, optimization, testing, and benchmarking of various DL-based model architectures, in pursuit of finding the optimal parameter set for semantic pixel-level FOI classification for the domain of diamond growth. To develop and test different DL-based model architectures mentioned, we used tensorflow [19], a deep learning library with strong visualization capabilities and dynamic, near-exhaustive options for use in deep learning and computer vision-driven model development. The segmentation models were trained on a machine with 32GB RAM, a 12-core 3.50 GHz processor, and a NVidia GeForce RTX 2080 Ti graphics card with 12GB VRAM. Our research on the model architecture development is performed across three state of the art semantic-segmentation model architectures using the image-label datasets procured from the SP3 module.

For our FCN-based implementation, we utilized MobileNet-V2 as the encoder backbone with 276M trainable parameters, outperforming Resnet and Xception variants. DeeplabV3 and DeeplabV3plus models utilized a modified Xception backbone with around 57M and 54M trainable parameters, respectively. All models were trained for 30–45 epochs with a batch size of 20 and a learning rate between $6 \times 10^{-6}$ and $3 \times 10^{-4}$. For loss optimization, we experimented with various functions including sparse categorical cross entropy and focal loss [20]. Focal loss proved more effective in handling imbalanced data distributions. In selecting evaluation metrics, we experimented with pixel accuracy (PA), mean pixel accuracy (MPA), intersection over union (IoU), and mean intersection over union (mIoU) [2]. While PA and MPA provide limited insights, IoU and mIoU offer better performance evaluation, with mIoU being our metric of choice due to its unbiased representation for datasets with class imbalances.

Module development consists of two main steps. The first involves obtaining a baseline model with 80% classification accuracy, iteratively improved through model-assisted labeling. The second step focuses on refining the model for remaining image-label datasets to achieve benchmark FOI classification accuracies for the diamond growth domain.

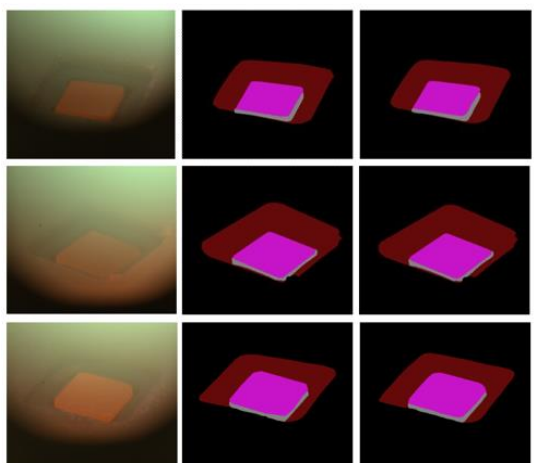

Figure 3. Visual depiction of segmentation and model results on unseen test data; Input Image, Actual/Labeled FOI mask, Predicted FOI mask from left to right; Data presented in black, red, pink, and green represent the different independent FOIs; background, pocket-holder, diamond top and diamond sides, respectively.

Step 1: Developing the baseline model - This step builds on the research elaborated on as part of Data Labeling Experiment Module, SP3. and Figure 2 to create an effective baseline model for use in the model assisted labeling (MAL) process. This enables the labelers to correct the model-assisted labels overlayed on image batches instead of having to start annotating each image from scratch. The labeled image batches are used to train the baseline model iteratively until a threshold accuracy detailed previous module.

An essential aspect of MAL baseline model development is the implementation of selective augmentation learning (SAL). During the evaluation phase, images with low segmentation accuracies undergo selective augmentation using predefined data augmentation techniques [30]. Our augmentation suite includes linear transformations like rotation and resizing, as well as nonlinear transformations like noise addition and JPEG compression. These transformations simulate unpredictable variations in images without needing real-world samples. For example, composite rotation and shear transformations mimic camera vibrations, while embossing, sharpening, blurring, and adding gaussian noise simulate optical hardware distortions. Augmented image datasets are then used to re-train the baseline model iteratively. The re-training loop continues for 5 iterations, with low performing images indicating potential data integrity issues. A re-labeling report identifies images for correction by the labeling team. This iterative process repeats until a baseline model surpassing desired accuracy benchmarks (>80%) is achieved. The finalized baseline model is then employed in the MAL process for subsequent image batches, advancing to the next step for the final segmentation model development.

Step 2: Developing the final model - The second step, as depicted in Figure 4, employs the baseline model developed in Step 1 to pre-annotate all subsequent batches of images for use in development of the final segmentation model. The labeling team corrects the pre-annotated labels in line with steps outlined in previous section. The entire loop between the image labeling, selective augmentations, re-training and re-labeling through custom reports is repeated until a final model with desired threshold accuracy benchmarks (>95%) is obtained. This final model is then saved to the model database for use in production. Figure 3 graphically depicts results of a successful segmentation model implementation developed using our pipeline.

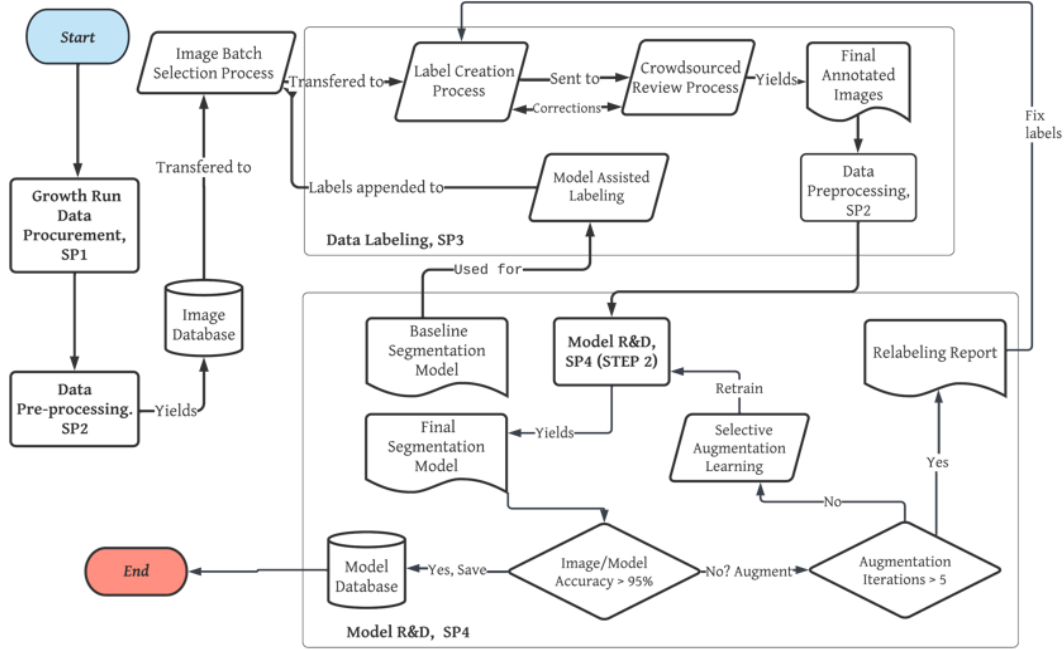

Figure 4. Flow for Step 2 of Model Research and Development Module

**Model Results-Post Analytics**

The model post analytics module employs the infrastructure and insights gained from SP4 to re-train new model versions based on key hyper-parametric factors of interest like 1) DL model architecture, 2) input image resolution, and 3) Dataset size variations. Overall, we trained 18 different models for the broader semantic segmentation objective with above-mentioned hyper-parameter variations. The results of our experiments are detailed in Table I and elaborated over in subsequent sub-sections.

TABLE I. FOI-level segmentation IoU based accuracy on unseen test data.

|  |  | 256x256 image size | | | 512x512 image size | | |
| --- | --- | --- | --- | --- | --- | --- | --- |
| Models | Dataset size | pocket-holder (IoU) | diamond-top (IoU) | diamond-sides (IoU) | pocket-holder (IoU) | diamond-top (IoU) | diamond-sides (IoU) |
| FCN | 10000 | 92.7 | 96.33 | 76.8 | 92.54 | 96.83 | 82.28 |
| DeeplabV3 | 10000 | 94.09 | 96.43 | 79.496 | 95.62 | 97.89 | 85.77 |
| DeeplabV3+ | 10000 | 95.09 | 97.70 | 85.67 | 96.31 | 98.60 | 91.64 |
| FCN | 5000 | 93.05 | 96.37 | 75.64 | 91.08 | 96.10 | 76.52 |
| DeeplabV3 | 5000 | 93.26 | 96.27 | 79.85 | 94.11 | 97.20 | 82.72 |
| DeeplabV3+ | 5000 | 94.48 | 97.67 | 86.22 | 95.54 | 98.23 | 88.36 |
| FCN | 2000 | 91.43 | 95.83 | 74.40 | 88.74 | 93.07 | 68.20 |
| DeeplabV3 | 2000 | 92.45 | 95.61 | 74.45 | 92.72 | 96.76 | 74.26 |
| DeeplabV3+ | 2000 | 92.58 | 96.51 | 79.53 | 93.02 | 96.82 | 80.37 |

**DL model architecture**. In this section, we discuss results of our experiments on understanding the effect of variations in model architectures over the validation accuracy on unseen image data. Based on our research in Background Section, we employed 3 model architectures for this category of experiments - FCN, DeepLabV3 and DeepLabV3Plus. The results of these experiments are detailed in Table I. Generally, our experiments determined DeeplabV3plus as the best performing model architecture on input data of all resolutions. However, it performed worse than other variants when using a lower dataset size. This requirement for a relatively larger dataset size can be attributed to the model architecture's larger complexity in terms of trainable parameters and the general distribution of the architecture compared to FCN and DeeplabV3. The mIoU based per performance of the FCN architecture is the lowest owing to lesser complexity (ability to capture complex pixel distribution patterns) and significant loss of information in the feature embedding output of the encoder due to pooling operations. DeeplabV3 performs better than FCN due to the use of atrouous convolutional operations contributing to richer object boundaries and thereby a higher segmentation accuracy.

**Input Image Resolution Variations.** In this section, we discuss results of our experiments on understanding the effect of variations in input dataset resolution over the validation accuracy on unseen image data. We employed 2 input image resolutions for this category of experiments - 256x256 and 512x512. The results of these experiments are detailed in Table I.

We observed the best accuracy measures when higher resolutions are used, which can be attributed to the richness of pixel-level features available for higher precision in learnt representations for FOI contours. Table I indicates a performance gain with an increase in dataset size when using higher resolution images. Lower resolution images, however, experience only marginal improvement in performance with an increase in dataset size.

**Dataset Size Variations.** In this section, we discuss results of our experiments on understanding the effect of variations in dataset size over the validation accuracy on unseen image data. We employed 3 data augmentation rates for this category of experiments - 2x, 5x and 10x. The results of these experiments are detailed in Table I.

The general trend observed across all model architectures and resolution of images used is that an increase in dataset size contributes to an increase in accuracy. We intend to experiment on this further as part of future research with bigger dataset sizes and augmentation rates.

**Consolidated Results**. Table I details the independent FOI-wise and overall segmentation performance breakdown of the models developed across all the hyper-parameter variations mentioned in the introduction to this section. Across all the experiments, we obtained maximum FOI-specific accuracies of 96.31% for the pocket-holder, 98.60% for the diamond top and 91.64% for the diamond side. The slight discrepancy in accuracy metrics for the diamond side can be attributed to the number of labeled instances for diamond sides being lower than the other FOIs. This is in turn due to the visible diamond side boundary progressively disappearing during the later phases of the diamond growth process.

Our best segmentation accuracy metrics were achieved on the model trained using the DeeplabV3plus architecture with a dataset size of 10,000 and input image resolution of $512 \times 512$. Using an unseen test dataset of 60 images, our best performing model achieved a class level IoU of 96.31%, 98.60% and 91.64% for the pocket holder, diamond top and diamond side FOIs respectively with an overall mean IOU of 95.43%. Also, our results on the best performing model show that our pipeline generalizes well on unseen test data even in low dataset volume environments. Table I provides a comprehensive overview of the accuracy benchmark established and of the insights gained from our model development experiments.

**Outlook and future research**

We described in this paper, how we successfully developed and evaluated a novel feature annotation and extraction pipeline from diamond growth data based on deep learning over a low-volume high-complexity dataset environment. This approach achieved state-of-the-art accuracy metrics in classifying and predicting FOIs contours over a semantic image segmentation objective. Additionally, as part of our pipeline, we propose and implement a data development architecture that significantly reduces the time and cost involved and increases the data integrity of the image annotation process. Across all the experiments, we obtained maximum feature of interest-specific accuracies of 96.31% for the pocket-holder, 98.60% for the diamond top, and 91.64% for the diamond side. Additionally, we successfully performed experiments to ascertain key hyper-parameters contributing to optimal precision in our segmentation outputs. For low-volume data experiments like those in crystal synthesis, these results are encouraging and can lead to more wide-spread implementation of deep learning approaches to crystal process development.

As part of future research efforts, we will further test the efficiency of our pipeline by developing more growth run datasets with higher complexity of spatial features in the training data. Additionally, as mentioned in Background Section, components developed for the feature extraction pipeline will be used as a starting template in development of the defect extraction pipeline that comprises development of object detection and segmentation models to classify and isolate accurate pixel masks of defects like polycrystalline growth, center defects and edge defects. Additionally, results from the feature and defect extraction models will be used to guide our research on the frame prediction pipeline that predicts future image states of diamond growth from 2 hrs to 16 hrs into the future. The feature and defect extraction models, when applied to the resulting predicted output images from the frame prediction model, will be pivotal in forming automated hypotheses over the effect of reactor parameters like temperature and pressure on defects and feature characteristics. These aspects will be tackled as part of future research to be conducted this year.